\documentclass[10pt,twocolumn,letterpaper]{article}

\usepackage{iccv}              

\usepackage{graphicx}
\usepackage{amsmath}
\usepackage{amssymb}
\usepackage{booktabs}
\usepackage{multirow}
\usepackage{color}
\usepackage{bbm}
\usepackage{tabu}
\usepackage{threeparttable}
\usepackage{wrapfig}
\usepackage{xcolor}
\usepackage{colortbl}

\definecolor{iccvblue}{rgb}{0.21,0.49,0.74}
\usepackage[pagebackref,breaklinks,colorlinks,allcolors=iccvblue]{hyperref}

\def\paperID{8956} 
\def\confName{ICCV}
\def\confYear{2025}

\title{Probabilistic Prototype Calibration of Vision-Language Models for \\ Generalized Few-shot Semantic Segmentation}

\author{Jie Liu$^{1*}$, Jiayi Shen$^{1}$, Pan Zhou$^{2}$\thanks{Corresponding authors.}~, Jan-Jakob Sonke$^{3}$, Efstratios Gavves$^1$\\
{\normalsize{$^1$University of Amsterdam}}~~~{\normalsize{$^2$Singapore Management University}}~~~{\normalsize{$^3$The Netherlands Cancer Institute}}}

\begin{document}
\maketitle

\begin{abstract}
Generalized Few-Shot Semantic Segmentation (GFSS) aims to extend a segmentation model to novel classes with only a few annotated examples while maintaining performance on base classes.
Recently, pretrained vision-language models (VLMs) such as CLIP have been leveraged in GFSS to improve generalization on novel classes through multi-modal prototypes learning.
However, existing prototype-based methods are inherently deterministic, limiting the adaptability of learned prototypes to diverse samples, particularly for novel classes with scarce annotations.
To address this, we propose FewCLIP, a probabilistic prototype calibration framework over multi-modal prototypes from the pretrained CLIP, thus providing more adaptive prototype learning for GFSS.
Specifically, FewCLIP first introduces a prototype calibration mechanism, which refines frozen textual prototypes with learnable visual calibration prototypes, leading to a more discriminative and adaptive representation.
Furthermore, unlike deterministic prototype learning techniques, FewCLIP introduces distribution regularization over these calibration prototypes.
This probabilistic formulation ensures structured and uncertainty-aware prototype learning, effectively mitigating overfitting to limited novel class data while enhancing generalization.
Extensive experimental results on PASCAL-5$^i$ and COCO-20$^i$ datasets demonstrate that our proposed FewCLIP significantly outperforms state-of-the-art approaches across both GFSS and class-incremental setting.
The code is available at \url{https://github.com/jliu4ai/FewCLIP}.

\end{abstract}    
\section{Introduction}
\label{sec:intro}
Semantic segmentation~\cite{chen2017deeplab, PSP_CVPR_2017, cheng2022masked} aims to assign semantic labels to each pixel of an image and has achieved remarkable success in the deep learning era. However, traditional semantic segmentation heavily relies on vast amounts of data with pixel-wise annotations and often struggles with classes absent from the training set, a.k.a. novel classes.  To tackle these challenges, few-shot semantic segmentation (FSS)~\cite{CANet_CVPR_2019, PANet_ICCV_2019, PFENet_TPAMI_2020, cermelli2021prototype, yang2020prototype, Dong_BMVC_2018, liu2022intermediate, liu2022dynamic, min2021hypercorrelation, zhang2019pyramid, peng2023hierarchical} has emerged. FSS first trains a base model on many annotated data of base classes to learn general knowledge, and then fine-tunes the model on a few annotated samples of novel classes for adaptation. However, FSS primarily focuses on the binary segmentation of novel classes at test time, but lacks identification ability on base classes and multi-novel classes~\cite{liu2023learning,huang2023prototypical}.

\begin{figure}[!t]
	\centering
	\includegraphics[width=\linewidth]{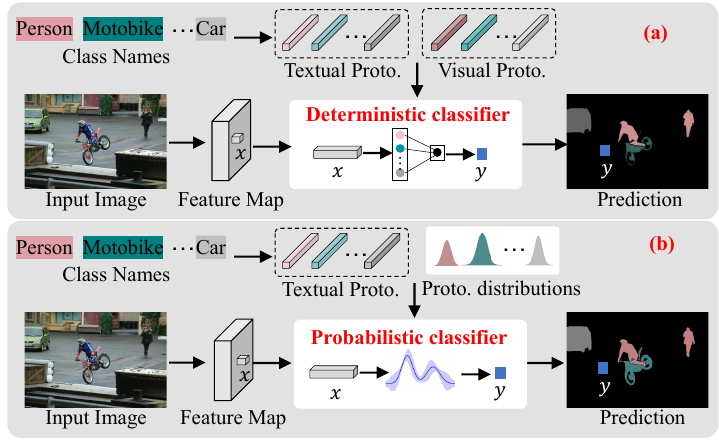}
	\vspace{-5mm}
	\caption{\textbf{Concept diagram comparing deterministic and probabilistic prototype learning with pre-trained CLIP in GFSS task.} (a) Existing methods integrate a learnable or precomputed visual prototype with a frozen textual prototype for each class to perform pixel-wise deterministic classification.
	(b) Our proposed FewCLIP models the class-wise visual calibration prototype as a distribution, capturing intra-class variations and uncertainty. By representing prototypes with a probabilistic distribution, our method mitigates prototype overfitting to scarce data, thereby improving generalization to novel classes.
	}
	\label{fig:motivation}
	\vspace{-6mm}
\end{figure}

As a more practical and ambitious extension, generalized few-shot semantic segmentation (GFSS)~\cite{tian2022generalized, liu2023learning,huang2023prototypical,myers2021generalized, hajimiri2023strong,hossain2024visual,sakai2025surprisingly,wang2024rethinking} aims to enable the segmentation of both base and novel classes within a given image. 
Typically, GFSS consists of two phases: (1) the \textit{base class training phase}, where a base segmentation model is trained on base classes with abundant annotated samples, and (2) the \textit{novel class registration phase}, where novel classes are integrated into the model using only a few annotated samples per class.
Due to the significant data availability gap between these two phases, the primary challenge in GFSS lies in \emph{enhancing model generalization to novel classes while preserving performance on base classes}.
Generally, existing methods~\cite{huang2023prototypical,liu2023learning,yang2023multi} tackle this challenge by decoupling the prototype learning of base and novel classes, ensuring that only novel class prototypes are optimized during the novel class registration phase.
More recently, O3S~\cite{yang2023multi} introduced multi-modal prototypes learning with pre-trained CLIP~\cite{radford2021learning} to further improve model generalization on novel classes.
However, as shown in Fig.~\ref{fig:motivation} (a), existing prototype-based methods remain deterministic, failing to account for the inherent uncertainty in learned visual prototypes.
For instance, images from the same class can exhibit significant variations in appearance, pose, and orientation, making it challenging for a deterministic prototype to adequately capture all possible variations.
As a result, overlooking uncertainties in prototype learning can lead to overfitting on the limited training samples and causing the model to forget  generalizable knowledge~\cite{kumar2022fine,jha2024clap4clip}. 
This issue is particularly severe during the novel class registration phase, where visual prototypes must be learned from only a few annotated samples, further hindering model generalization.

\noindent{\textbf{Contributions.}} To address these limitations, we propose FewCLIP, a novel probabilistic prototype calibration  framework that integrates uncertainty-aware prototype learning with pretrained vision-language models (VLMs).  
To enhance generalization to novel classes, we first introduce a prototype calibration mechanism, where FewCLIP calibrates frozen textual prototypes from pre-trained CLIP with learnable visual calibration prototypes in a class-specific manner.
This mechanism preserves the generalizable knowledge from CLIP while incorporating class-specific information by learning calibration prototypes from supervised data. 
By optimizing only the visual prototypes for novel classes during the novel class registration phase, FewCLIP ensures effective adaptation to novel classes while preserving base class knowledge.
Furthermore, to mitigate the overfitting of calibration prototypes on limited observed samples—especially in the novel class registration phase, we introduce probabilistic prototype calibration. 
As illustrated in Fig.~\ref{fig:motivation} (b), instead of representing calibration prototypes as deterministic vectors, we model them as continuous probability distributions, allowing them to capture uncertainty and intra-class variation in the learned representations.
This probabilistic formulation provides a distribution regularization over visual calibration prototypes, mitigating overfitting to limited novel class samples, thereby improving model generalization over these classes. 
In summary, our contributions are threefold:

$\bullet$ We introduce a prototype calibration mechanism for GFSS, adapting frozen textual prototypes with visual calibration prototypes to enhance model generalization to novel classes while maintaining performance on base classes. 

$\bullet$ We propose FewCLIP, a probabilistic prototype calibration framework modeling calibration prototypes as probability distributions to mitigate prototype overfitting and enable uncertainty-aware prototype learning.

$\bullet$ Extensive experiments on PASCAL-5$^i$ and COCO-20$^i$ benchmarks demonstrate the superiority of our method in both GFSS and class-incremental FSS settings, achieving around 10\% mIoU improvement on novel classes.

\section{Related Work}
\label{sec:rw}
\noindent\textbf{Generalized Few-shot Semantic Segmentation (GFSS).} Unlike conventional few-shot semantic segmentation, GFSS~\cite{tian2022generalized} aims to recognize both base and novel classes in an image, leveraging sufficient data from base classes and a few data from novel classes. 
One line of research in GFSS seeks to \textit{mimic} the novel class registration during base class training, employing techniques such as contextual aggregation~\cite{tian2022generalized}, orthogonality~\cite{liu2023learning}, and kernel-based learning~\cite{huang2023prototypical}.  
These methods enable a seamless extension of the label space when incorporating novel classes. 
Another research direction explores various fine-tuning~\cite{myers2021generalized, hajimiri2023strong,hossain2024visual} strategy during novel class registration to effectively adapt models to novel classes. 
More recently, BCM~\cite{sakai2025surprisingly} offer a strong baseline for GFSS by training novel-class models by supervised learning. 
O3S~\cite{yang2023multi} and BCM~\cite{wang2024rethinking} investigate how multi-modal knowledge from vision-language models can enhance model generalization to novel classes.

\noindent\textbf{Vision-Language Models (VLMs).}
Pre-trained Vision-Language Models (VLMs)~\cite{jia2021scaling,li2019visualbert,radford2021learning} connect the image representation with text embedding and have made rapid progress in recent years.  Contrastive Language-Image Pre-training (CLIP)~\cite{radford2021learning} is one of the most popular VLMs, by aligning the representation  of two modalities in the same embedding space via contrastive learning. 
In practice, CLIP has demonstrated strong zero-shot generalization across various downstream tasks, including visual question answering~\cite{jiang2022finetuning}, image retrieval~\cite{liu2021image,baldrati2022effective}, and dense prediction~\cite{zhou2023zegclip,rao2022denseclip,liu2023clip,wang2022cris}. 
Recent studies~\cite{yang2023multi,wang2024rethinking} have also explored leveraging CLIP’s cross-modal knowledge to improve GFSS. However, without task-specific fine-tuning, its performance on dense prediction tasks remains limited.
In this work, we investigate effective task-specific finetuning strategies for CLIP to address challenges in GFSS.
\begin{figure*}[!t]
	\vspace{-4mm}
	\centering
	\includegraphics[width=0.95\linewidth]{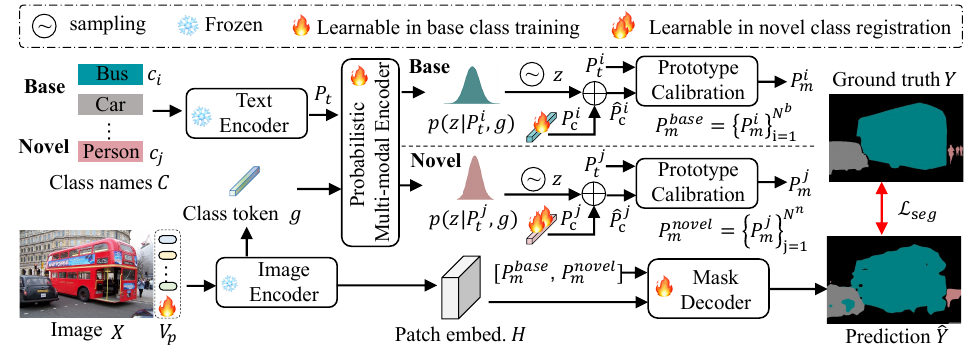}
	\vspace{-4mm}
	\caption{\textbf{ Overview of the proposed FewCLIP with probabilistic prototype calibration.} Given an image and a set of class names, we first extract textual prototypes $P_t$ and visual embeddings [$g, H$] using CLIP's text and image encoders, respectively. 
		For each class, we learn a sample-specific distribution \(p(z_i|P_t^i, g)\) and a calibration prototype \(P_c^i\), which are further integrated into a probabilistic calibration prototype \(\hat{P}_c^i\).
		Next, the probabilistic calibration prototype \(\hat{P}_c^i\) is used to calibrate the textual prototype \(P_t^i\).
		Finally, the calibrated prototype \(P_m\) and patch embeddings \(H\) are fed into a mask decoder to generate the final mask prediction.
		Novel classes can be seamlessly integrated into the model by fine-tuning a calibration prototype \(P_c^j\) with few annotated data in the novel class registration phase.
		\vspace{-4mm}}
	\label{fig2}
\end{figure*} 

\noindent\textbf{Parameter Efficient Fine-Tuning of VLMs.}
Without task-specific fine-tuning, VLMs often perform poorly on downstream out-of-domain data~\cite{jha2024clap4clip}. For such case, parameter-efficient fine-tuning (PEFT) methods have been developed. One approach is to learn lightweight feature adapter modules for the textual and/or visual pathways~\cite{zhou2023zegclip,wang2022cris}. Another strategy is soft prompt tuning, which introduces a small set of continuous tokens as inputs to the frozen visual and/or textual encoders to capture task-specific information~\cite{rao2022denseclip,liu2023clip, jia2022visual}.
To enhance fine-tuning with uncertainty modeling in downstream tasks, some methods adopt probabilistic fine-tuning techniques, including variational prompt tuning~\cite{derakhshani2023bayesian}, probabilistic prompt learning~\cite{kwon2023probabilisticpromptlearningdense}, and continual learning with probabilistic fine-tuning~\cite{jha2024clap4clip}.
In this work, we introduce a probabilistic prototype calibration mechanism for VLMs to improve GFSS performance on novel classes without compromising base class.

\section{Preliminaries}
\label{sec:problem}
\noindent\textbf{Generalized Few-shot Segmentation (GFSS).}
GFSS aims to extend the label space of a pre-trained segmentation model to novel classes using only a few annotated training samples while maintaining strong performance on base classes. Typically, GFSS follows a two-phase training scheme~\cite{myers2021generalized,tian2022generalized, liu2023learning}:

$\bullet$ In the \emph{base class training phase}, a base segmentation model is trained on the base classes $C^b=\{c_i\}_{i=1}^{N^b}$, where $N^b$ is the number of base classes. Each base class has a sufficient number of annotated training images, enabling the model to learn general and transferable representations. 

$\bullet$ In the \emph{novel class registration phase}, novel classes $C^n=\{c_{i}\}_{i=N^b+1}^{N^b+N^n}$ are incorporated to the pre-trained base model, where $N^n$ is the number of novel classes. This is typically achieved through fine-tuning, as each novel class has only a few annotated training samples, e.g., $K$ samples. 

In the \emph{inference phase}, given a test image, the trained model performs semantic segmentation by classifying each pixel into one of the $1+N^b+N^n$ classes, including $N^b$ base classes, $N^n$ novel classes, and a background class.
However, existing GFSS methods suffer from poor performance on either novel classes, as seen in prototype-based fine-tuning approaches~\cite{huang2023prototypical,liu2023learning,yang2023multi}, or base classes, as in adapter-based fine-tuning methods~\cite{myers2021generalized,jia2022visual}.  
Therefore, the key challenge in GFSS is to \emph{improving the generalization on novel classes} with limited annotated data while \emph{maintaining performance on base classes}~\cite{tian2022generalized,  hajimiri2023strong,liu2023learning}.  

\noindent\textbf{CLIP Fine-tuning for Segmentation Tasks.}
Typically, recent studies~\cite{zhou2023zegclip, zhang2024exploring, hoyer2024semivl, zhou2024unlocking} freeze the text and image encoders of pre-trained CLIP in the segmentation tasks.  
To effectively adapt CLIP for segmentation, these works incorporate visual prompt tuning~\cite{zhou2022learning, jia2022visual} into the image encoder and learn a mask decoder to enhance text-image matching.
Specifically, given \(N\) class names \(\{c_i\}_{i=1}^N\), we encode them using the CLIP text encoder $\mathcal{F}_{\text{text}}$ to obtain the corresponding  set of textual prototype representations, denoted as $P_t=\{P_t^i\}_{i=1}^{N}$, where each prototype is computed as:
\begin{equation}
		P_t^i= \mathcal{F}_{\text{text}}(c_i).
\end{equation} 
Similarly, for a given image \(X\), we extract its visual representation using the frozen image encoder \(\mathcal{F}_{\text{image}}\):
\begin{equation}
 [g, H] = \mathcal{F}_{\text{image}}([X, V_p]),
\end{equation} 
where $g\in\mathbb{R}^{d} $ denotes the class token embedding with size \(d\), and $H=\{h_i\}_{i=1}^{L}\in\mathbb{R}^{L\times d}$ denotes $L$ local visual feature patch \(h_i\).
\(V_p\) represents a set of deep visual prompts~\citep{zhou2022learning}, which introduce task-specific learnable parameters into the input space.
Then, the segmentation mask is computed as:
\begin{equation}
		\hat{Y} = \mathcal{F}_{\text{mask}}(\psi(P_t),\Psi(H)),
\label{eq:mask}
\end{equation} 
where \(\mathcal{F}_{\text{mask}}\) is a learnable mask decoder deployed as a lightweight transformer. \(\psi\) and \(\Psi\) denote the projection layers on the prototypes and feature patches, respectively.

\section{Methodology}
\label{sec:md}

\subsection{Overview}
Fig.~\ref{fig2} illustrates the overview of our proposed FewCLIP with probabilistic prototype calibration for Generalized Few-shot Semantic Segmentation (GFSS).
Given an image \(X\) and a set of class names \(C\), our goal is to accurately segment images from both base and novel classes  using the general knowledge in Vision-Language Models (VLMs).
First, the image encoder \(\mathcal{F}_{\text{image}}\) and text encoder \(\mathcal{F}_{\text{text}}\) respectively extract visual embeddings [$g, H$] and textual prototypes \(P_t\), where $g$ and $H$ are respectively the class token and patch tokens. 
Next, the class token \(g\) and textual prototypes \(P_t\) are fed into a probabilistic multi-modal encoder to generate a sample-specific distribution for each class, e.g., \(p(z|P_t^i, g)\) for base class \(c_i\). 
The sampled representations from this distribution are then fused with a learnable calibration prototype \(P_c^i\) to obtain the probabilistic calibration prototype \(\hat{P}_c^i\), which is further used to calibrate corresponding textual prototype \(P_t^i\).
Novel classes can be seamlessly integrated into the model by learning a calibration prototype for each class during novel class registration.
Finally, the calibrated multi-modal prototype \(P_m\) from both base and novel classes, along with the patch embeddings \(H\), are fed into a mask decoder \(\mathcal{F}_{\text{mask}}\) to predict the mask \(\hat{Y}\).

\subsection{Prototype Calibration}
With the pre-trained CLIP model, our goal is to learn a compact multi-modal prototype for each base and novel class, enabling accurate segmentation of both during inference.
While textual prototypes derived from class names provide strong generalizable information, but textual prototypes alone lack fine-grained visual semantics~\citep{yang2023multi,rao2022denseclip,zhou2023zegclip}, often leading to suboptimal segmentation performance.
To mitigate this issue, we introduce a set of learnable calibration prototypes \(P_c=\{P_c^i\}_{i=1}^{N}\) to calibrate the frozen textual prototypes \(P_t=\{P_t^i\}_{i=1}^{N}\) by incorporating class-specific visual information from observed annotated samples.
Here,  $N$ dynamically changes according to the training stage of GFSS, i.e., $N= N^b$ during the base class training phase and $N= N^b+N^n$ during the novel class registration phase. 
Formally, the prototype calibration is defined as:
\begin{equation}
    P_m = P_t \odot P_c + P_t \in \mathbb{R}^{N\times d}
    \label{eq:correct}
\end{equation}
where $\odot$ denotes the Hadamard product and $P_m$ represents the set of calibrated multi-modal prototypes. The term $P_t \odot P_c$ provides a multi-modal corrective bias to the frozen textual prototypes, yielding a more compact and generalizable multi-modal prototype. Experimental results  in Table~\ref{table_p_format} of Sec.~\ref{sec:ablation} show superiority   of the calibration format~\eqref{eq:correct}. Next, we concatenate $P_m$ with the background prototype $P_0$ to obtain the final calibrated prototypes $    \hat{P}_m$ as follows:
\begin{equation}
    \hat{P}_m = \phi_\text{proj}([P_0, P_m]) \in \mathbb{R}^{(N+1)\times d},
    \label{eq:cs}
\end{equation}
where $\phi_\text{proj}(\cdot)$ is a two-layered MLPs. The background prototype $P_0$ is learnable in the base class training phase but remain frozen in the novel class registration phase.

This decoupled   calibration allows our model to retain CLIP’s generalization capacity within textual prototypes while acquiring class-specific visual semantics through calibration prototypes. During novel class registration, we freeze all parameters from the base class training phase, including base class calibration prototypes \(P_c^\text{base}\), and only fine-tune  the novel class prototypes \(P_c^\text{novel}\). This  seamlessly integrates novel classes without affecting base class representations. By replacing \(P_t\) with \(\hat{P}_m\) in Eq.~(\ref{eq:mask}), we obtain the predicted segmentation mask \(\hat{M}\). Therefore, our decoupled calibration  enhances generalization on novel classes while preserving strong performance on base classes.


\subsection{Probabilistic Prototype Calibration}
The novel class registration phase in GFSS suffers from data scarcity, making it challenging to learn a deterministic and representative calibration prototype for each novel class.
On one hand, the limited number of available samples can lead to overfitting, resulting in poor generalization to unseen data.  
On the other hand, high intra-class variance—arising from differences in appearance, shape, and texture—necessitates a prototype that can explicitly capture uncertainty in its representation.
To address these challenges, we introduce probabilistic prototype calibration to  model the calibration prototype as a probability distribution rather than a deterministic prototype vector. 
This probabilistic formulation can better capture intra-class variations while mitigating overfitting in the few-shot setting.

\noindent\textbf{Probabilistic Modeling.}
For each class \(c_i\), we define a sample-specific probability distribution \(p(z|P_t^i, g)\), conditioned on the class-specific textual prototype \(P_t^i\) and sample-specific class token \(g\) extracted from the input image. 
Here we assume that \(p(z|P_t^i, g)\) follows a Gaussian with a mean vector \(\mathbf{\mu}_i\) and a variance vector \(\mathbf{\sigma}_i\):
\begin{equation}
    p(z|P_t^i, g) \sim \mathcal{N}(\mathbf{\mu}_i, \mathbf{\sigma}_i).
\end{equation}
As shown in Fig.~\ref{fig2}, we design a probabilistic multi-modal encoder to infer the parameters of this distribution.  
Specifically, we employ a multi-head cross-attention layer ($\texttt{MHCA}$)  to process the textual prototype \(P_t^i\) and the visual class token \(g\), enabling effective cross-modal feature fusion.  
Here, \(P_t^i\) serves as the query, while both key and value vectors are derived from \(g\).  
The mean and variance vectors are then estimated via two separate two-layered MLPs \(\phi_\mu\) and \(\phi_\sigma\):  
\begin{align}
	\mathbf{\mu}_i = \phi_\mu(\texttt{MHCA}(P_t^i, g)),  \quad 
	\mathbf{\sigma}_i = \phi_\sigma(\texttt{MHCA}(P_t^i, g)).
\end{align} 
To further enhance adaptability, we model the calibration prototype distribution across all classes as  a Mixture of Gaussian distributions (MoG):  
\begin{equation}
	p(z|P_t, g) \sim \sum_{i=1}^N\mathcal{N}(\mathbf{\mu}_i, \mathbf{\sigma}_i).
\end{equation}  
This formulation enables diverse prototype sampling, enhancing adaptability to novel classes while preserving class consistency for robust segmentation.
Given \(p(z|P_t, g)\), we sample \(M\) latent prototypes for each class, namely, \(z =\{z^1, \cdots, z^M\}\sim p(z|P_t, g)\).
This sampling follows the reparameterization trick~\cite{kingma2015variational}, ensuring differentiability for efficient optimization.
Accordingly, these sampled calibration prototypes represent learned distributions over class representations, capturing the inherent variability in visual appearance within each class.

\noindent\textbf{Probabilistic Calibration.} 
Given the sampled calibration prototypes \(z\), we compute the final calibration prototype by combining them with the learnable calibration prototype \(P_c\):  
\begin{equation}
	\hat{P}_c=\{P_c+z^m\}_{m=1}^M.
\end{equation}  
This formulation incorporates both deterministic components from learnable calibration prototype \(P_c\) and probabilistic variations from the sampled latent prototypes \(z\).
The deterministic part captures visual semantics from the limited annotated samples, while the probabilistic component \(z\) acts as a regularization mechanism, preventing overfitting and improving generalization to novel instances.
By substituting the \(P_c\) in Eq.~(\ref{eq:correct}) with \(\hat{P}_c\), we perform probabilistic prototype calibration and segmentation,  yielding \(M\) mask predictions.
Finally, we adopt pixel-wise mean of these predictions as final mask, while the pixel-wise variance quantifies the model’s epistemic uncertainty, offering a measure of confidence in the segmentation results.

\subsection{Training Objective}
Our training objective consists of two parts, a cross-entropy loss that supervises pixel-wise segmentation accuracy,  and a KL divergence regularization term that aligns the distribution of calibration prototypes with a Gaussian prior \(\mathcal{N}(\mathbf{0},\mathbf{I})\), ensuring smooth and well-structured prototype learning~\cite{jha2024clap4clip,kwon2023probabilisticpromptlearningdense}. Formally, the overall  loss is defined as:
\begin{equation} \mathcal{L} = \mathcal{L}_{\text{CE}}(\hat{Y},Y) + \lambda \mathcal{L}_\text{KL}(p(z | P_t, g) || \mathcal{N}(\mathbf{0}, \mathbf{I})), \end{equation}
where \(\lambda\) is a  balancing weight and set as 0.001 during training. 
During base class training, we freeze the image and text encoders of the pre-trained CLIP model, optimizing the visual prompts \(V_p\), the probabilistic multi-modal encoder, the calibration prototypes, and the mask decoder to adapt CLIP to the segmentation task.
In the novel class registration phase, we freeze all model parameters and only update the calibration prototypes \(P_c\) for novel classes, allowing the model to accommodate new categories without affecting previously learned base class representations.

\subsection{Extension to class incremental setting}
\label{sec:CIFSS}
We further extend our model to the class-incremental setting, a.k.a., Class-Incremental Few-Shot Semantic Segmentation (CIFSS)~\cite{huang2023prototypical}, where novel classes are introduced sequentially over multiple sessions. In CIFSS, training begins with a base session \(S_b\), which has sufficient labeled data. This is followed by a sequence of novel sessions \(S_n = \{S_n^1, \cdots, S_n^T\}\), each containing a limited number of annotated samples. Importantly, at any given time step \(t\), we only have access to the data from session \(S_n^t\), and there is no overlap between novel sessions.  

Our model naturally extends to this setting by incrementally updating the text and calibration prototypes. Formally, let \(P_t^s\) and \(P_c^s\) denote the text prototype and calibration prototype of session \(s\), respectively. When transitioning to the next session \(s+1\), we update these prototypes as follows:  
\begin{align}  
	P_t^{s+1} &= [P_t^s, P_t^{s+1}] \in \mathbb{R}^{N^{s+1} \times d}, \notag \\  
	P_c^{s+1} &= [P_c^s, P_c^{s+1}] \in \mathbb{R}^{N^{s+1} \times d},  
\end{align}  
where [$\cdot$,$\cdot$] is the concatenation operation, and $N^{s+1}$ is the number of classes in the session $s+1$. During incremental learning,  we exclusively register novel classes by optimizing the calibration prototypes $P_c^{s+1}$ using limited data from these novel classes.
\section{Experiments}
\label{sec:exp}
\noindent\textbf{Datasets.}
We test our method on two GFSS benchmarks: PASCAL-5$^i$~\cite{everingham2010pascal_voc, shaban2017one} and COCO-20$^i$~\cite{lin2014microsoft, shaban2017one}. 
PASCAL-5$^i$, a subset of PASCAL VOC 2012~\cite{everingham2010pascal_voc}, 
consists of 12,301 images with high-quality, pixel-wise annotation across 20 semantic categories, divided into 4 splits for cross-validation (5 classes per split).  COCO-20$^i$, a subset of MSCOCO~\cite{lin2014microsoft}, has 122,218 images annotated across 80 object classes which are divided into 4 splits for cross-validation (20 classes per split)~\cite{tian2022generalized}. In the experiment, classes from a specific split are set novel classes, while the remaining are base classes. For CIFSS, we follow \cite{huang2023prototypical} and adopt the same fold division with GFSS. Novel classes on both PASCAL-5$^i$ and COCO-20$^i$ are divided into 5 incremental sessions. 

\noindent\textbf{Evaluation metrics.} 
We use the mean intersection-over-union (mIoU) as the performance metric. Specifically, we report mIoU for base and novel classes as \textbf{mIoU$_\mathcal{B}$} and \textbf{mIoU$_\mathcal{N}$}, respectively. We also report the overall mIoU on all classes, i.e., \textbf{mIoU$_\mathcal{O}$}. As base classes is the majority,  mIoU$_\mathcal{O}$ could predominantly reflect mIoU$_\mathcal{B}$~\cite{baek2021exploiting}.
So we follow~\cite{huang2023prototypical,liu2023learning,hajimiri2023strong} and adopt the harmonic mean of mIoU$_\mathcal{B}$ and mIoU$_\mathcal{N}$, denoted as \textbf{$h$IoU}, as the primary metric.

\begin{table*}[!t]
\Large
\caption{\textbf{Quantitative results under 1-shot and 5-shot GFSS settings on PASCAL-5$^i$ and COCO-20$^i$}. $\mathcal{I}$ and $\mathcal{T}$ denote image and text are provided for novel class registration, respectively. The harmonic mean $\emph{h}\text{IoU}$ is employed as the primary metric to balance the significance of the results for base classes, $\emph{m}\text{IoU}_{\mathcal{B}}$ and novel class, $\emph{m}\text{IoU}_{\mathcal{N}}$. Best-performing results are highlighted in bold. Our method FewCLIP demonstrates notable improvements, particularly in the 1-shot setting.}
\vspace{-2mm}
\begin{threeparttable}
\resizebox{1.0\linewidth}{!}{
\begin{tabular}{lc|cccc|cccc||cccc|cccc}
\toprule

\multirow{3}{*}{\bf Methods} & \multirow{3}{*}{\bf Info.} & \multicolumn{8}{c||}{\bf PASCAL-5$^i$} & \multicolumn{8}{c}{\bf COCO-20$^i$}\\

\cmidrule(lr){3-10}  \cmidrule(lr){11-18} & & \multicolumn{4}{c|}{\bf 1-shot} & \multicolumn{4}{c||}{\bf 5-shot} & \multicolumn{4}{c|}{\bf 1-shot} & \multicolumn{4}{c}{\bf 5-shot}\\ 
\cmidrule(l){3-6} \cmidrule(l){7-10} \cmidrule(l){11-14} \cmidrule(l){15-18}
& & \bf $\emph{m}\text{IoU}_{\mathcal{B}}$ & \bf $\emph{m}\text{IoU}_{\mathcal{N}}$ & \bf $\emph{m}\text{IoU}_{\mathcal{O}}$ & \bf $\emph{h}\text{IoU}$ & \bf $\emph{m}\text{IoU}_{\mathcal{B}}$ & \bf $\emph{m}\text{IoU}_{\mathcal{N}}$ & \bf $\emph{m}\text{IoU}_{\mathcal{O}}$ & \bf  $\emph{h}\text{IoU}$ & \bf $\emph{m}\text{IoU}_{\mathcal{B}}$ & \bf $\emph{m}\text{IoU}_{\mathcal{N}}$ & \bf $\emph{m}\text{IoU}_{\mathcal{O}}$ & \bf $\emph{h}\text{IoU}$ & \bf $\emph{m}\text{IoU}_{\mathcal{B}}$ & \bf $\emph{m}\text{IoU}_{\mathcal{N}}$ & \bf $\emph{m}\text{IoU}_{\mathcal{O}}$ & \bf  $\emph{h}\text{IoU}$\\ 
\midrule
BAM \cite{lang2022learning} & $\mathcal{I}$ & 64.65 & 17.06 & 53.32 & 27.00 & 65.28 & 19.99 & 54.50 & 30.61 & 44.08 & 6.74 & 34.86 & 11.69 & 46.10 & 11.21 & 37.49 & 18.03 \\
CAPL  \cite{tian2022generalized} & $\mathcal{I}$ & 65.48 & 18.85 & 54.38 & 29.27 & 66.14 & 22.41 & 55.72 & 31.96 & 44.61 & 7.05 & 35.46 & 12.18 & 45.24 & 11.05 & 36.80 & 17.76\\ 
PKL \cite{huang2023prototypical} & $\mathcal{I}$ & 68.84 &  26.90 &  58.86 &  37.83 & 69.22 &  34.40 &  61.18  &  45.42  & 46.36 &  11.04 & 37.71 & 17.83 & 46.77 & 14.91 &  38.90 &  22.61 \\
DIaM~\cite{hajimiri2023strong} & $\mathcal{I}$ & 70.89  & 35.11  & - & 46.96 & 70.85 & 55.31  & -  &  63.08 & 48.28  & 17.22  & - & 25.39 & 48.37 & 28.73  & -  & 36.05 \\
PoP~\cite{liu2023learning} & $\mathcal{I}$ & 73.92  & 35.51  & 64.77 & 47.97 & 74.78 & 55.87  & 70.28  & 63.96 & \bf 54.71  & 15.31  & \bf 44.98 & 23.92 & \bf 54.90 & 29.97  & \bf 48.75 & \bf 38.77 \\ 
BCM~\cite{sakai2025surprisingly} & $\mathcal{I}$ &  71.15 & 41.24  & 56.20 & 52.22 & 71.23 & 55.36  & 63.29  & 62.30 & 49.43  & 18.28  & 33.85 & 26.69 & 49.88  & 30.16  & 40.24  &  37.59 \\
\midrule
VPT \cite{jia2022visual} & $\mathcal{I}$+$\mathcal{T}$ & 55.93 & 47.74 & 55.43 & 51.43 & 61.38 & 52.76 & 59.40 & 56.63 & 43.40 & 24.10 & 39.14 & 30.77 & 44.81 & 25.76 & 39.20 & 32.69\\ 
O3S~\cite{yang2023multi} & $\mathcal{I}$+$\mathcal{T}$ & 71.20  & 39.44  & - & 50.76 & 72.22  & 57.53  & - & 64.04 & 45.83  & 19.48  & - & 27.34 & - & -  & - & - \\
\rowcolor[gray]{.9} \bf FewCLIP  & $\mathcal{I}$+$\mathcal{T}$ & \bf 76.68 & \bf 52.70 & \bf 71.72 & \bf 62.03  & \bf 76.31 & \bf 59.21 & \bf 73.06  & \bf 66.44 & 48.82 & \bf 26.79  & 43.83 & \bf 34.60  & 49.05 & \bf 31.52 & 45.17  & 38.38 \\ 
\bottomrule
\end{tabular}}
\end{threeparttable} 
\label{table_GFSS}
\end{table*}
\begin{figure}[!t]
\centering
\includegraphics[width=\columnwidth]{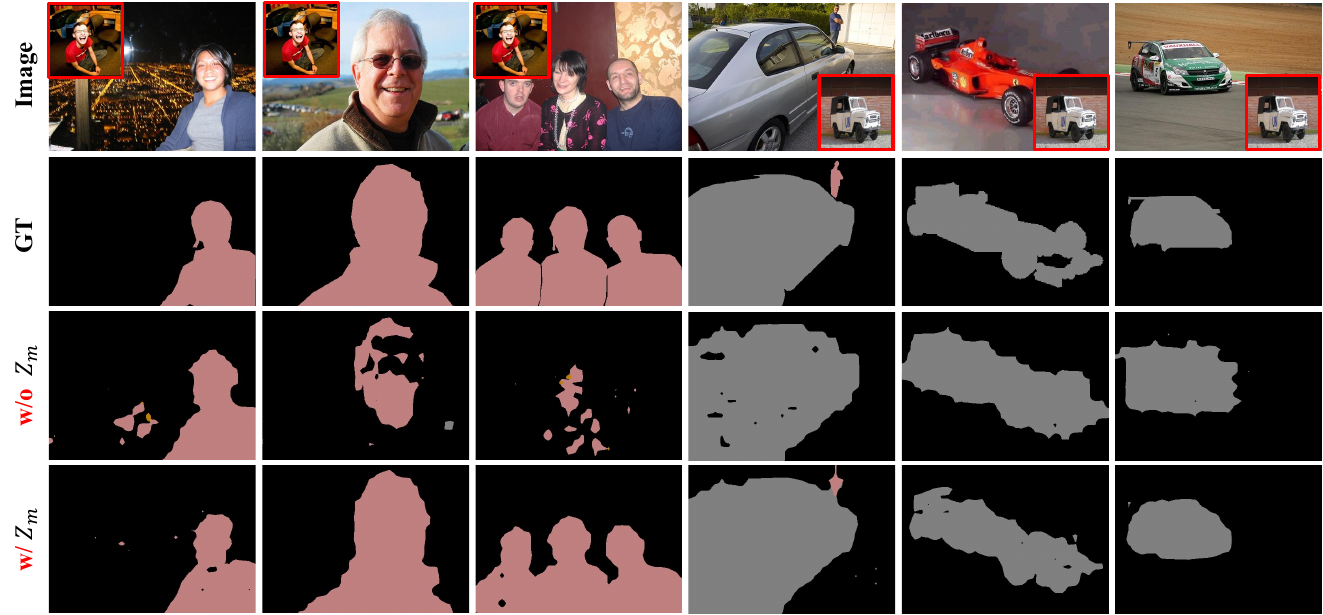}
\caption{\textbf{Benefit of introducing latent prototypes \(z_m\) from the prototype distribution when dealing with large object variations.} The image masked with red rectangles denotes the training sample. Left triplet: \colorbox{pink!80}{Person}, right triplet: \colorbox{gray!50}{Car}.}
\label{fig:pir_vis}
\vspace{-2mm}
\end{figure}
\noindent\textbf{Implementation details.}
We use the pre-trained CLIP ViT-B/16 model with input image and mask resolutions of 512x512. During the base class learning phase, we use AdamW with a learning rate of 2.5e-4, a weight decay parameter of 1e-2, and a batch size of 8. Then we train PASCAL-5$^i$/COCO-20$^i$ for 100/50 epochs. In the novel class registration phase, we adopt SGD with a learning rate of 0.5  to train the calibration prototypes for 100 epochs by using novel classes samples. See more details in Appendix.

\subsection{Comparison with state-of-the-art}
\noindent\textbf{GFSS setting.}
Table~\ref{table_GFSS} reports the quantitative results for 1-shot and 5-shot settings  on PASCAL-5$^i$ and COCO-20$^i$. 
We compare our method against two types of methods: visual prototype methods with only class-specific images are provided, and multi-modal prototype methods with both visual and textual information.
Table~\ref{table_GFSS} shows that visual prototype methods like PoP~\cite{liu2023learning} and BCM~\cite{sakai2025surprisingly} have much inferior performance on novel classes compared to base classes.
This performance gap arises because novel class prototypes, learned from only a few samples, struggle to generalize effectively to unseen data. 
In contrast, our FewCLIP consistently outperforms visual prototype methods, with particularly notable improvements on novel classes in the 1-shot scenario. For instance, in the 1-shot scenario, our FewCLIP outperforms BCM~\cite{sakai2025surprisingly} by a margin of 11.54$\%$ and 8.51$\%$ in $m\text{IoU}_\mathcal{N}$ on PASCAL-5$^i$ and COCO-20$^i$, respectively.
The results indicate the advantage of introducing textual prototype from pre-trained CLIP model and our novel prototype calibration mechanism in improving model generalization on novel classes in GFSS task. 

Multi-modal prototype methods such as VPT~\cite{jia2022visual} and O3S~\cite{yang2023multi} adopt both textual and visual information from CLIP.
Compared with these methods, our FewCLIP not only preserves performance on base classes but also achieves significant improvements on novel classes. Specifically, in terms of $\emph{h}\text{IoU}$, our FewCLIP surpasses VPT by 10.60\% in the 1-shot setting and by 9.81\% in the 5-shot setting on the PASCAL-5$^i$ dataset.
The result shows the impact of the probabilistic prototype calibration module in FewCLIP.
Moreover, our FewCLIP outperforms O3S by a large margin, e.g., 11.27\% in 1-shot setting on the PASCAL-5$^i$ dataset, and 7.26\% in 1-shot setting on the COCO-20$^i$ dataset. 
The results further demonstrate the effectiveness of our probabilistic prototype calibration method in improving model generalization in GFSS task.
\begin{table}[tp]
\Large
\centering 
\setlength{\abovecaptionskip}{0cm}
\setlength{\belowcaptionskip}{0cm}
\caption{\textbf{Component analysis of FewCLIP}. PC denotes prototype calibration. Results for 1-shot and 5-shot settings are reported.}
\vspace{2mm}
\begin{threeparttable}
\resizebox{\linewidth}{!}{
\begin{tabular}{lcccccc}
\toprule
\multirow{2}{*}{\bf Method} & \multicolumn{3}{c}{\bf 1-shot} & \multicolumn{3}{c}{\bf 5-shot} \\ 
\cmidrule(l){2-4} \cmidrule(l){5-7}
 & \bf $\emph{m}\text{IoU}_{\mathcal{B}}$ & \bf $\emph{m}\text{IoU}_{\mathcal{N}}$ & \bf $\emph{h}\text{IoU}$ & \bf $\emph{m}\text{IoU}_{\mathcal{B}}$ & \bf $\emph{m}\text{IoU}_{\mathcal{N}}$ & \bf $\emph{h}\text{IoU}$ \cr
  \midrule
 w/o PC  & 72.57 & 33.75 & 46.02 & 72.57 & 33.75 & 46.02 \cr
 Deterministic PC & 75.35 & 51.36  &  60.79 &  75.10 &  59.07 & 65.22 \cr
\rowcolor[gray]{.9}Probabilistic PC & \bf 76.68 & \bf 52.70  & \bf 62.03   & \bf 76.31 & \bf 59.21 & \bf 66.44 \cr
\bottomrule
\end{tabular}}
\end{threeparttable} 
\label{table_component}
\vspace{-2mm}
\end{table}


\begin{table*}[!t]
\setlength{\abovecaptionskip}{0cm}
\setlength{\belowcaptionskip}{0cm}
\caption{\textbf{Quantitative results under the 1-shot CIFSS setting on the PASCAL-5$^i$ and COCO-20$^i$}. All baseline results are quoted from \cite{huang2023prototypical}. The harmonic mean $\emph{h}\text{IoU}$ is employed as the primary metric to balance the significance of the results for base classes, $\emph{m}\text{IoU}_{\mathcal{B}}$ and novel class, $\emph{m}\text{IoU}_{\mathcal{N}}$. The best-performing results are highlighted in bold. The proposed model FewCLIP exhibits markedly superior performance than previous methods across all incremental sessions.}
\vspace{2mm}
\resizebox{\linewidth}{!}{
\begin{tabular}{c|l|c|ccc|ccc|ccc|ccc|ccc}
\toprule
& \multirow{2}{*}{Methods} & sesson 0 & \multicolumn{3}{c|}{session 1} & \multicolumn{3}{c|}{session 2} & \multicolumn{3}{c|}{session 3} & \multicolumn{3}{c|}{session 4} & \multicolumn{3}{c}{session 5} \\ \cline{3-18} 
& & $\emph{m}\text{IoU}_{\mathcal{B}}$ & $\emph{m}\text{IoU}_{\mathcal{B}}$ & $\emph{m}\text{IoU}_{\mathcal{N}}$ & $\emph{h}\text{IoU}$ & $\emph{m}\text{IoU}_{\mathcal{B}}$ & $\emph{m}\text{IoU}_{\mathcal{N}}$ & $\emph{h}\text{IoU}$ & $\emph{m}\text{IoU}_{\mathcal{B}}$ & $\emph{m}\text{IoU}_{\mathcal{N}}$ & $\emph{h}\text{IoU}$ & $\emph{m}\text{IoU}_{\mathcal{B}}$ & $\emph{m}\text{IoU}_{\mathcal{N}}$ & $\emph{h}\text{IoU}$ & $\emph{m}\text{IoU}_{\mathcal{B}}$ & $\emph{m}\text{IoU}_{\mathcal{N}}$ & $\emph{h}\text{IoU}$ \\ \hline
\multirow{6}{*}{\rotatebox{90}{PASCAL-5$^i$}} & PFENet \cite{PFENet_TPAMI_2020} & 74.43 & 67.21 & 10.34 & 17.92 & 66.62 & 14.12 & 23.30 & 64.97 & 11.12 & 18.99 & 65.77 & 12.08 & 20.41 & 62.50 & 11.08 & 18.82\\
& iFS-RCNN \cite{nguyen2022ifs} & 72.43 & 67.71 & 11.08 & 19.04 & 67.17 & 16.63 & 26.66 & 65.51 & 12.94 & 21.61 & 65.48 & 11.75 & 19.92 & 63.36 & 12.39 & 20.73\\
& CAPL \cite{tian2022generalized} & 74.86 & 70.51 & 17.94 & 28.60 & 69.07 & 21.11 & 32.34 & 67.89 & 18.03 & 28.49 & 67.14 & 19.57 & 30.31 & 65.53 & 19.08 & 29.55 \\
& PIFS \cite{cermelli2021prototype} & 75.04 & 71.16 & 19.49 & 30.60 & 69.80 & 24.33 & 36.08 & 68.05 & 19.92 & 30.82 & 68.44 & 21.08 & 32.23 & 65.83 & 20.99 & 31.83\\
& PKL \cite{huang2023prototypical} &  75.49 & 72.93 &  24.60 &  36.79 &  71.21 &  29.32 &  41.54 &  70.76 &  25.75 &  37.76 &  69.98 &  26.11 &  37.88 &  68.40 & 27.28 & 39.00 \\
\rowcolor[gray]{.9} \multicolumn{1}{c|}{\cellcolor{white}} &  \bf FewCLIP & \bf 79.39 & \bf 76.48 & \bf 39.63 & \bf 52.21 & \bf 76.20 & \bf 49.81 & \bf 60.24 & \bf 75.82 & \bf 56.32 & \bf 64.63 & \bf 75.80 & \bf 54.27 & \bf 63.25 & \bf 75.44 & \bf 51.62 & \bf 61.30 \\ \hline \hline
\multirow{7}{*}{\rotatebox{90}{COCO-20$^i$}} & PFENet \cite{PFENet_TPAMI_2020}& 54.11 & 45.79 & 4.33 & 7.91 & 44.58 & 5.18 & 9.28 & 43.07 & 3.36 & 6.23 & 42.40 & 5.64 & 9.96 & 41.99 & 5.05 & 9.02\\
 & iFS-RCNN \cite{nguyen2022ifs} & 53.42 & 50.43 & 5.03 & 9.15 & 45.52 & 6.60 & 11.53 & 43.86 & 5.07 & 9.09 & 43.37 & 7.11 & 12.22 & 42.83 & 6.63 & 11.48\\
 & CAPL \cite{tian2022generalized} & 54.43 & 50.25 & 6.69 & 11.81 & 46.67 & 7.04 & 12.23 & 45.92 & 5.53 & 9.87 & 44.08 & 7.77 & 13.21 & 43.85 & 7.51 & 12.82 \\
 & BAM \cite{lang2022learning} &  54.80 & \bf 52.04 & 8.83 & 15.10 & 49.01 & 8.94 & 15.12 & 47.03 & 7.03 & 12.23 & 45.56 & 9.19 & 15.29 & 44.48 & 8.83 & 14.73\\
 & PIFS \cite{cermelli2021prototype} & 54.27 & 51.59 & 8.90 & 15.18 & 48.52 & 9.53 & 15.93 & 46.96 & 6.94 & 12.09 & 45.69 & 9.93 & 16.31 & 44.27 & 8.80 & 14.68 \\
 & PKL \cite{huang2023prototypical} & 54.39 & 51.78 & 11.47 &  18.55 & 48.33 & 11.22 & 18.08 & 47.28 &  9.61 & 15.89 & 46.25 & 11.47 & 18.33 & 45.24 & 10.74 & 17.31 \\
\rowcolor[gray]{.9} \multicolumn{1}{c|}{\cellcolor{white}} & \bf FewCLIP & \bf 55.99  &  49.19 & \bf 33.52 & \bf 39.87 & \bf 49.17 & \bf 36.41 & \bf 41.84 & \bf 48.93 & \bf 29.88 & \bf 37.10 & \bf 49.16 & \bf 29.09 & \bf 36.55 & \bf 48.79 & \bf 30.46 & \bf 37.51 \\
\bottomrule
\end{tabular}}
\label{table_CIFSS}
\vspace{-2mm}
\end{table*}

\noindent\textbf{CIFSS setting.} Table~\ref{table_CIFSS} shows the quantitative results for the CIFSS setting on the PASCAL-5$^i$ and COCO-20$^i$. Overall, our method exhibits significantly better performance than existing approaches in all three metrics, \textit{i.e.}, $\emph{m}\text{IoU}_{\mathcal{B}}$, $\emph{m}\text{IoU}_{\mathcal{N}}$, and $\emph{h}\text{IoU}$. For instance, in session 1, our method exceeds the PKL~\cite{huang2023prototypical} by 15.42$\%$ and 21.32$\%$ in $\emph{h}\text{IoU}$ on PASCAL-5$^i$ and COCO-20$^i$, respectively. As the novel class session increases, the superiority of our method in handling both base and novel classes becomes even more pronounced. The results generally embodies the effectiveness of our model in dynamically accommodating novel classes in a session stream.
\begin{table}[tp]
\Large
  \centering  
  \setlength{\abovecaptionskip}{0cm}
  \setlength{\belowcaptionskip}{0cm}
  \caption{\textbf{Effects of prototype calibration formats in class-specific calibration}. $P_t$ and $P_c$ denote frozen textual embeddings and learnable corrective prototypes. [$\cdot$, $\cdot$] stands for concatenation.}
  \vspace{2mm}
  \begin{threeparttable}
  \resizebox{1.02\linewidth}{!}{
  \begin{tabular}{lcccccc}
  \toprule
  \multirow{2}{*}{\bf Formats}  & \multicolumn{3}{c}{\bf 1-shot} & \multicolumn{3}{c}{\bf 5-shot} \\ 
  \cmidrule(l){2-4} \cmidrule(l){5-7}
   & \bf $\emph{m}\text{IoU}_{\mathcal{B}}$ & \bf $\emph{m}\text{IoU}_{\mathcal{N}}$& \bf $\emph{h}\text{IoU}$ & \bf $\emph{m}\text{IoU}_{\mathcal{B}}$ & \bf $\emph{m}\text{IoU}_{\mathcal{N}}$ & \bf $\emph{h}\text{IoU}$ \cr
  \midrule
$ P_t+P_c$ & 73.43  & 32.25  & 44.76 & 70.05 & 48.62 & 57.16 \cr
$ P_t-P_c$ & 73.81  & 32.42  & 45.01 & 70.50 & 47.50 & 56.46 \cr
$ P_t*P_c$ & 70.26  & 31.69  & 42.35 & 72.29 & 42.23 & 52.73 \cr
$[P_t, P_c]$ & 71.41  & 27.64  & 38.92 & 68.43 & 43.35 & 52.70 \cr
$[P_t*P_c, P_t]$ & 66.28  & 24.30  & 35.50 & 61.34 & 37.96 & 46.40 \cr
\rowcolor[gray]{.9} $P_t*P_c+P_t$& \bf 76.68 & \bf 52.70  & \bf 62.03   & \bf 76.31 & \bf 59.21 & \bf 66.44 \cr
\bottomrule
\end{tabular}}
\end{threeparttable} 
\label{table_p_format}
\vspace{-2mm}
\end{table}

\begin{figure*}[!t]
\centering
\includegraphics[width=0.95\linewidth]{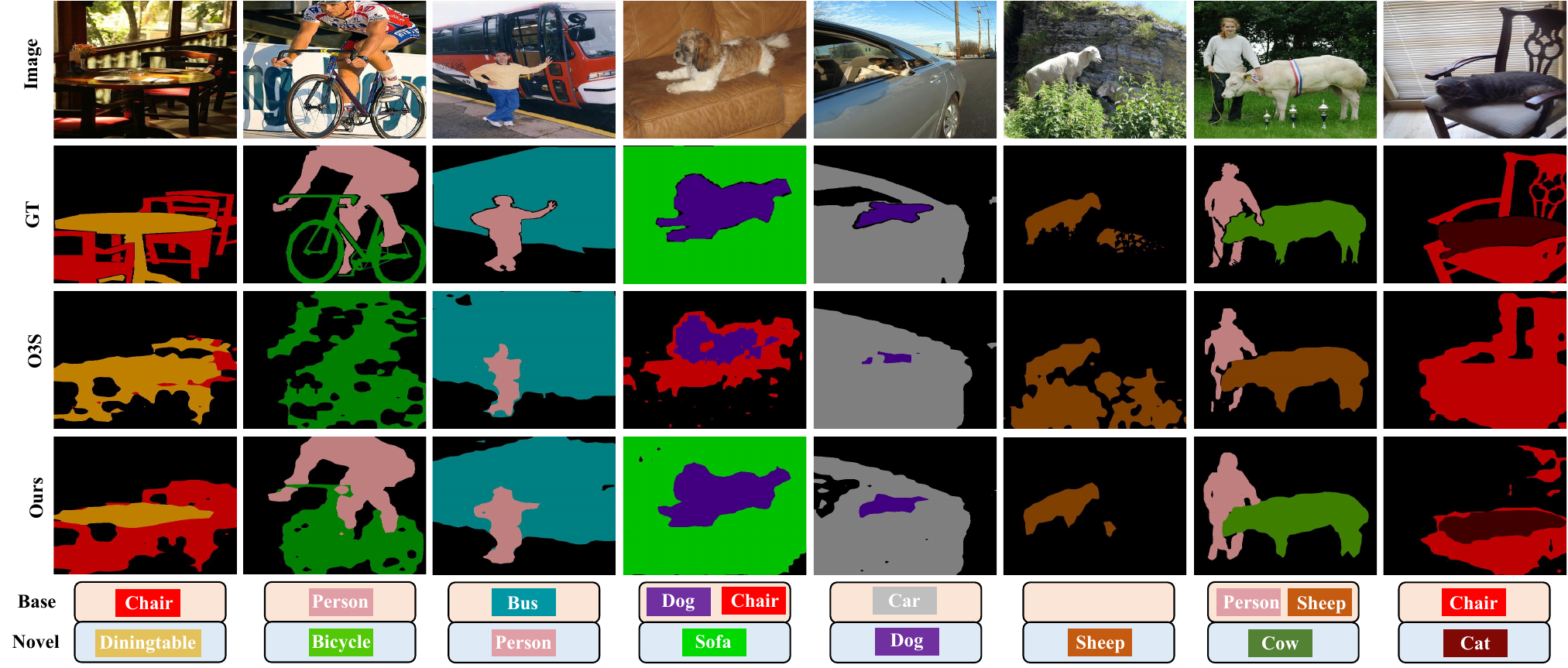}
\caption{\textbf{Qualitative examples of O3S~\cite{yang2023multi} and our method on PASCAL-5$^i$}. While O3S 'forget' base classes (2-th column) or fail to generalize to novel classes (7-th column), our model effectively generalizes to novel classes without compromising base classes.}
\label{fig:vis}
\vspace{-2mm}
\end{figure*}

\noindent\textbf{Qualitative analysis.}
Fig.~\ref{fig:vis} showcases eight visual examples of O3S~\cite{yang2023multi} and our FewCLIP on PASCAL-5$^i$ under the GFSS setting. By comparison, our FewCLIP achieves more accurate segmentation results on both base and novel classes. 
For instance, as shown in the first two columns,  O3S detects novel classes \texttt{Diningtable} and \texttt{Bicycle} but fails to recognize base classes \texttt{Chair} and \texttt{Person}, while our method precisely identifies both base and novel classes. More qualitative results are available in Appendix.

\subsection{Ablation Studies}
\label{sec:ablation}
We analyze the impact of two key components in our model: the prototype calibration mechanism and the probabilistic prototype calibration mechanism
To establish a baseline, we use a model that follows a similar fusion strategy as O3S~\cite{yang2023multi}, integrating textual and visual prototypes without explicit prototype calibration.
As shown in Table~\ref{table_component}, this baseline exhibits limited generalization to novel classes, highlighting the challenge of learning reliable visual prototypes with scarce annotated samples.
Introducing our deterministic prototype calibration significantly boosts performance on novel classes, improving $\emph{m}\text{IoU}_{\mathcal{N}}$ from 33.75\% to 51.36\%, demonstrating its effectiveness in enhancing multi-modal representation learning.
Furthermore, the probabilistic prototype calibration further refines performance on both base and novel classes by regularizing the learned prototypes and mitigating overfitting.
Fig.~\ref{fig:pir_vis} provides a qualitative analysis, showcasing its ability to address large intra-class variations and improve model generalization.
\begin{table}[tp]
\Large
\centering  
\setlength{\abovecaptionskip}{0cm}
\setlength{\belowcaptionskip}{0cm}
\caption{\textbf{Benefits of provided information during novel class registration.} By integrating visual and text information simultaneously, our method achieves best results.}
\vspace{2mm}
\begin{threeparttable}
\resizebox{1.02\linewidth}{!}{
\begin{tabular}{cccccccc}
\toprule
\multirow{2}{*}{\bf Vision}  & \multirow{2}{*}{\bf Text} & \multicolumn{3}{c}{\bf 1-shot} & \multicolumn{3}{c}{\bf 5-shot} \\ 
\cmidrule(l){3-5} \cmidrule(l){6-8}
&  & \bf $\emph{m}\text{IoU}_{\mathcal{B}}$ & \bf $\emph{m}\text{IoU}_{\mathcal{N}}$& \bf $\emph{h}\text{IoU}$ & \bf $\emph{m}\text{IoU}_{\mathcal{B}}$ & \bf $\emph{m}\text{IoU}_{\mathcal{N}}$ & \bf $\emph{h}\text{IoU}$ \cr
\midrule
\checkmark & - & 75.45 & 31.67  & 44.29 & 73.69 & 52.26 & 60.86\ \cr
- & \checkmark  & 72.57 & 33.75 & 46.02 & 72.57 & 33.75 & 46.02 \cr
\rowcolor[gray]{.9} \checkmark & \checkmark & \bf 76.68 & \bf 52.70  & \bf 62.03   & \bf 76.31 & \bf 59.21 & \bf 66.44 \cr
\bottomrule
\end{tabular}}
\end{threeparttable} 
\label{table_inform}
\vspace{-3mm}
\end{table}

\noindent\textbf{Effects of prototype calibration formats.}
To investigate the effects of different prototype calibration formats on improving model generalization, 
we explore six different prototype calibration formats.   
As shown in Table~\ref{table_p_format}, 
the format $P_t*P_c+P_t$ employed in our module consistently outperforms alternative strategies across all evaluation metrics. Notably, operations such as \texttt{add}, \texttt{difference}, and \texttt{dot-product} demonstrate superior performance in enhancing novel class recognition compared to \texttt{concatenation}. \texttt{concatenation} usually performs poorly on both base and novel classes due to its tendency to weaken class-specific signals.
Consequently, $P_t*P_c+P_t$ was selected as the prototype calibration format in the proposed class-specific calibration.

\noindent\textbf{Effects of visual and textual information.}
Our method requires both image (visual) and class names (textual) inputs during the novel class registration phase.
To show the benefits of visual and textual information within our model, we investigate the single-modal situation, namely, only images or class names are available. 
In such case, we derive prototypes for novel classes directly from a limited number of images or class names as final prototypes for mask prediction.
Table~\ref{table_inform} reveals that in the 1-shot setting, models using only class names outperform their counterparts that rely solely on images, particularly for novel classes. However, as the number of available images increases, the image-only model exhibits significant performance gains on the novel classes. 
These findings underscore the critical roles both images and class names play in enhancing the model's capacity for generalization to novel classes, particularly when a small amount of images are available.
\begin{table}[tp]
\Large
\centering  
\setlength{\abovecaptionskip}{0cm}
\setlength{\belowcaptionskip}{0cm}
\caption{\textbf{Benefits of different fine-tuning (FT) methods during novel class registration.} $P_t$ denotes textual embeddings, $[V_p, \mathcal{F}_\text{mask}, P_0]$ denotes parameters learned in base class learning, and $P_c$ denotes the introduced corrective prototypes.}
\vspace{2mm}
\begin{threeparttable}
\resizebox{1.02\linewidth}{!}{
\begin{tabular}{lcccccc}
\toprule
\multirow{2}{*}{\bf FT Methods}  & \multicolumn{3}{c}{\bf 1-shot} & \multicolumn{3}{c}{\bf 5-shot} \\ 
\cmidrule(l){2-4} \cmidrule(l){5-7}
& \bf $\emph{m}\text{IoU}_{\mathcal{B}}$ & \bf $\emph{m}\text{IoU}_{\mathcal{N}}$& \bf $\emph{h}\text{IoU}$ & \bf $\emph{m}\text{IoU}_{\mathcal{B}}$ & \bf $\emph{m}\text{IoU}_{\mathcal{N}}$ & \bf $\emph{h}\text{IoU}$ \cr
\midrule
w/o FT  & 72.57 & 33.75 & 46.02 & 72.57 & 33.75 & 46.02 \cr
FT $P_t$ & 73.67  & 35.09  & 46.83 & 70.58 & 48.50 & 57.16 \cr
FT $[V_p, \mathcal{F}_\text{mask}, P_0]$ &  55.93 & 47.74  & 51.43 & 61.38  & 52.76 & 59.40 \cr
\rowcolor[gray]{.9} FT $P_c$ & \bf 76.68 & \bf 52.70  & \bf 62.03   & \bf 76.31 & \bf 59.21 & \bf 66.44 \cr
\bottomrule
\end{tabular}}
\end{threeparttable} 
\label{table_finetune}
\vspace{-4mm}
\end{table}

\noindent\textbf{Effects of different fine-tuning strategies.}
We investigate the effects of different fine-tuning (FT) strategies in the novel class registration phase. Specifically, \textrm{w/o FT} denotes directly matching the frozen textual prototypes $P_t$ and image embedding $H$ to predict mask. $P_t$ denotes that we fine-tune the textual prototypes  on a few images from novel classes. $[V_p, \mathcal{F}_\text{mask}, P_0]$ refers to fine-tune the parameters of visual prompts \(V_P\), mask decoder \(\mathcal{F}_\text{mask}\), and background prototype \(P_0\) .
 Table~\ref{table_finetune} indicates that, 
\textrm{w/o FT} and fine-tuning $P_t$ exhibit comparable performance on both base and novel classes, highlighting the robust generalization capabilities of the pre-trained CLIP model. Fine-tuning $[V_p, \mathcal{F}_\text{mask}, P_0]$ notably improves novel class segmentation but at the cost of degraded base class performance, because this fine-tuning strategy affects the well-learned features for base classes. 
In contrast, the proposed method with fine-tuning calibration prototype $P_c$ achieves the best performance on novel classes without compromising base class performance. This highlights our model's exceptional ability to enhance generalization for novel classes while minimizing the impact on the performance of base classes.

\section{Conclusion}
\label{sec:con}
In this work, we propose FewCLIP, a novel probabilistic prototype calibration framework for GFSS that integrates prototype calibration and probabilistic modeling to enhance segmentation performance.
We first introduce a prototype calibration mechanism, which calibrates CLIP-derived textual prototypes with learnable visual prototypes, improving model generalization to novel classes while preserving base class performance. Additionally, to mitigates overfitting of visual prototypes, FewCLIP models visual prototypes as probability distributions, enabling uncertainty-aware prototype learning and improved generalization on novel classes. Extensive experiments and ablations on PASCAL-5$^i$ and COCO-20$^i$ validate FewCLIP’s effectiveness, demonstrating consistent improvements over state-of-the-art methods on both GFSS and CIFSS tasks.

\noindent\textbf{Limitations.}
FewCLIP relies on CLIP’s pretrained weights and class names, which may be suboptimal for domain-specific tasks where the visual and textual distributions deviate from CLIP’s training data. Future work could consider extend this work to the open-vocabulary segmentation task.

{
    \small
    \bibliographystyle{ieeenat_fullname}
    \bibliography{main}
}

\clearpage
\appendix


\begin{table*}[!t]
\Large
\setlength{\abovecaptionskip}{0cm}
\setlength{\belowcaptionskip}{0cm}
\captionsetup[]{width=\textwidth, justification=centering}
\caption{\textbf{Quantitative results for each fold of generalized few-shot semantic segmentation (GFSS) on PASCAL-5$^i$ benchmark}. $\dag$ denotes 1-shot results, while $\ddag$ represent 5-shot results.}
\resizebox{\textwidth}{!}{
\begin{tabular}{lcccccccccccccccc}
\toprule
\multirow{2}{*}{Methods} & \multicolumn{4}{c}{Fold-0} & \multicolumn{4}{c}{Fold-1} & \multicolumn{4}{c}{Fold-2} & \multicolumn{4}{c}{Fold-3} \\ 
\cmidrule(l){2-5} \cmidrule(l){6-9} \cmidrule(l){10-13}  \cmidrule(l){14-17}
 & $\emph{m}\text{IoU}_{\mathcal{B}}$ & $\emph{m}\text{IoU}_{\mathcal{N}}$ & $\emph{m}\text{IoU}_{\mathcal{O}}$ & $\emph{h}\text{IoU}$ & $\emph{m}\text{IoU}_{\mathcal{B}}$ & $\emph{m}\text{IoU}_{\mathcal{N}}$ & $\emph{m}\text{IoU}_{\mathcal{O}}$ & $\emph{h}\text{IoU}$ 
 & $\emph{m}\text{IoU}_{\mathcal{B}}$ & $\emph{m}\text{IoU}_{\mathcal{N}}$ & $\emph{m}\text{IoU}_{\mathcal{O}}$ & $\emph{h}\text{IoU}$
 & $\emph{m}\text{IoU}_{\mathcal{B}}$ & $\emph{m}\text{IoU}_{\mathcal{N}}$ & $\emph{m}\text{IoU}_{\mathcal{O}}$ & $\emph{h}\text{IoU}$
 \\ 
\midrule
BAM~\cite{lang2022learning}$\dag$  & 68.43 & 9.90 & 54.49 & 17.30 & 61.17 & 24.43 & 52.42 & 34.9  & 60.94 & 21.14 & 51.46 & 31.39 & 68.06 & 12.77 & 54.90 & 21.51 \\
CAPL~\cite{tian2022generalized}$\dag$ & 69.72 & 11.47 & 55.85 & 19.70 & 63.02 & 25.95 & 54.19 & 36.76 & 61.41 & 20.3 & 51.64 & 30.57 & 70.20 & 12.04 & 56.35 & 20.5 \\
PKL~\cite{prototypical}$\dag$ & 72.32 & 24.56 & 60.95 & 36.67 & 64.56 & 38.97 & 58.47 & 48.60 & 64.90 & 26.3 & 55.71 & 37.45 & 73.58 & 17.76 & 60.29 & 28.61  \\
DIaM~\cite{hajimiri2023strong}$\dag$ & 71.33 & 29.36 & - & 41.60 &69.54 &46.72 &- & 55.89 & 69.10 &27.07 & - & 38.90 &73.60 &37.30 &- & 49.51 \\
Baseline $\dag$ & 57.10  & 53.90  & 57.80 & 55.46 & 49.88 & 45.02  & 50.59  & 47.33 & 50.08 & 39.05  & 48.78  & 43.88 & 66.66 & 52.97  & 64.53  & 59.03 \\
 \rowcolor[gray]{.9} PPCN $\dag$ &78.88	&55.20	&73.90	&64.91	&74.72	&65.00	&73.30	&69.50	&73.78	&46.86	&68.11	&57.32	&79.41	&43.73	&71.58	&56.40  \\ \midrule

BAM~\cite{lang2022learning}$\ddag$  &67.73 &10.32 &54.01 &7.91 &62.45 &23.39 &53.15 &34.03 &60.84 &18.72 &50.81 &28.63 &68.18 &4.32 &52.98 &8.1 \\
CAPL~\cite{tian2022generalized}$\ddag$ &69.72 &11.47 &55.85 &19.70 &63.02 &25.95 &54.19 &36.76 &61.41 &20.3 &51.64 &30.57 &70.20 &12.04 &56.35 &20.5  \\
PKL~\cite{prototypical}$\ddag$ &72.47 &29.32 &62.20 &41.75 &67.25 &46.34 &62.28 &54.87 &63.37 &32.56 &56.03 &43.02 &73.77 &29.38 &63.20 &42.0 \\
DIaM~\cite{hajimiri2023strong}$\ddag$ &71.06 & 53.72 &- & 61.19 &69.63 & 63.33 &- & 66.33 & 69.12 & 54.01 &- & 60.64 & 73.60 & 50.19 &- & 59.68 \\
Baseline $\ddag$ & 68.62  &  56.43	& 66.78	& 61.93 & 54.67	& 54.93	& 56.48	& 54.8
& 54.01	& 44.81	& 48.89	& 48.98
& 68.22	& 54.87	& 65.43	& 60.82 \\
\rowcolor[gray]{.9} PPCN $\ddag$ &78.80	&63.10	&75.60	&60.50	&73.80	&68.22	&73.43	&70.91	&74.01	&55.24	&70.34	&63.27	&78.62	&51.29	&72.86	&62.08 \\
\bottomrule
\end{tabular}}
\label{pascal_fold}
\end{table*}
\begin{table*}[!t]
\Large
\setlength{\abovecaptionskip}{0cm}
\setlength{\belowcaptionskip}{0cm}
\caption{\textbf{Quantitative results for each fold of generalized few-shot semantic segmentation (GFSS) on COCO-20$^i$ benchmark.} $\dag$ denotes 1-shot results, while $\ddag$ represent 5-shot results.}
\resizebox{\linewidth}{!}{
\begin{tabular}{lcccccccccccccccc}
\toprule
\multirow{2}{*}{Methods} & \multicolumn{4}{c}{Fold-0} & \multicolumn{4}{c}{Fold-1} & \multicolumn{4}{c}{Fold-2} & \multicolumn{4}{c}{Fold-3} \\ 
\cmidrule(l){2-5} \cmidrule(l){6-9} \cmidrule(l){10-13}  \cmidrule(l){14-17}
 & $\emph{m}\text{IoU}_{\mathcal{B}}$ & $\emph{m}\text{IoU}_{\mathcal{N}}$ & $\emph{m}\text{IoU}_{\mathcal{O}}$ & $\emph{h}\text{IoU}$ & $\emph{m}\text{IoU}_{\mathcal{B}}$ & $\emph{m}\text{IoU}_{\mathcal{N}}$ & $\emph{m}\text{IoU}_{\mathcal{O}}$ & $\emph{h}\text{IoU}$ 
 & $\emph{m}\text{IoU}_{\mathcal{B}}$ & $\emph{m}\text{IoU}_{\mathcal{N}}$ & $\emph{m}\text{IoU}_{\mathcal{O}}$ & $\emph{h}\text{IoU}$
 & $\emph{m}\text{IoU}_{\mathcal{B}}$ & $\emph{m}\text{IoU}_{\mathcal{N}}$ & $\emph{m}\text{IoU}_{\mathcal{O}}$ & $\emph{h}\text{IoU}$
 \\ 
\midrule
BAM~\cite{lang2022learning} & 37.50 &3.39 &29.07 &6.15 &44.07 &8.84 &35.37 &14.73 &45.83 &4.41 &35.60 &8.05 &45.11 &7.70 &35.87 &13.15 \\
CAPL~\cite{tian2022generalized}$\dag$ &39.73 &5.26 &31.20 &9.29 &45.15 &10.12 &36.50 &16.53 &48.28 &6.82 &38.04 &11.95 &46.90 &8.36 &37.38 &14.10 \\
PKL~\cite{prototypical}$\dag$ &42.06 &8.43 &33.76 &14.04 &46.85 &12.89 &38.76 &20.22 &49.10 &10.40 &39.55 &17.16 &47.41 &12.44 &38.77 &19.71  \\
DIaM~\cite{hajimiri2023strong}$\dag$ & 49.01 &15.89 &- & 24.00 &46.83 & 19.50 &-  & 27.53 & 48.82 & 16.93 & - & 25.14 & 48.45 & 16.57 & - & 24.69 \\
Baseline $\dag$ &42.43	&21.22	&37.66	&28.29
&41.63	&24.5	&37.94	&30.05
&44.27	&24.61	&39.92	&31.64
&45.28	&26.07	&41.04	&33.09  \\
\rowcolor[gray]{.9} PPCN $\dag$ &45.77	&24.14	&40.86	&31.61	&48.15	&27.12	&43.43	&34.69	&51.7	&27.93	&46.26	&36.27	&49.67	&27.97	&44.77	&35.79  \\ \hline

BAM~\cite{lang2022learning}$\ddag$ &38.43 &4.73 &30.11 &8.42 &43.28 &8.59 &34.71 &14.33 &48.03 &6.31 &37.73 &11.15 &45.51 &7.84 &36.21 1&3.38 \\  
CAPL~\cite{tian2022generalized}$\ddag$  &40.14 &6.81 &31.91 &11.64 &43.79 &8.26 &35.02 &13.90 &45.71 &5.23 &35.71 &17.77 &46.39 &9.14 &37.19 &15.27 \\
PKL~\cite{prototypical}$\ddag$ &42.61 &10.79 &34.75 &16.82 &47.05 &16.77 &39.57 &24.73 &48.91 &15.88 &40.75 &23.96 &47.39 &16.95 &39.87 &24.97 \\
DIaM~\cite{hajimiri2023strong}$\ddag$ &48.90 &24.86 &- & 32.96 & 47.10 &33.94 - & 39.80 &49.12	&27.15 & - &34.97  &48.37 &28.95 &- & 36.22 \\
Baseline $\ddag$ &42.69	&23.87	&38.52	&30.62
&42.84	&24.95	&38.95	&31.54
&47.06	&27.59	&37.32	&34.78
&46.64	&26.61	&42.00	&33.82 \\
\rowcolor[gray]{.9} PPCN$\ddag$ & 45.9	&28.68	&42.1	&35.3	&48.41	&32.24	&44.88	&38.7	&51.65	&30.44	&46.84	&38.3	&50.22	&34.7	&46.84	&41.05 \\
\bottomrule
\end{tabular}}
\label{coco_fold}
\end{table*}

\section{Implementation Details}
\noindent\textbf{Dataset Setup.}
Following previous methods~\cite{liu2023learning,tian2022generalized,huang2023prototypical,hajimiri2023strong}, we adopt four-fold validation on PASCAL-5$^i$ and COCO-20$^i$ datasets for stable evaluation in generalized few-shot semantic segmentation (GFSS). Taking PASCAL-5$^i$ as an example, 20 target classes are evenly distributed into four folds. For a specific fold, its target classes are designated as novel classes, with limited annotated data available, such as in a 1-shot or 5-shot scenario. The target classes from the remaining folds form the base classes set, following~\cite{hajimiri2023strong}, we do not regard background as  one of base classes for more practical setting. Unlike the novel classes set, the base classes have abundant labeled data. For Class-Incremental Few-Shot Semantic Segmentation (CIFSS), we maintain four-fold cross-validation on PASCAL-5$^i$ and COCO-20$^i$ datasets using the same fold partitioning as in GFSS. In the selected fold, target classes are further divided into equal parts, termed as novel sessions. Simultaneously, all samples from base classes constitute the base session. These novel sessions are then incrementally processed in a stream.

\noindent\textbf{Model Optimization.}
The optimization of GFSS methods involves two phases: base class learning and novel class registration. In the base learning phase, the model is trained on base classes using extensive annotated data within a supervised framework. Following~\cite{liu2023learning,tian2022generalized,huang2023prototypical}, we exclude samples with novel classes during base class learning phase. During novel class registration, the model is further fine-tuned on novel classes using limited annotated data, typically 1 or 5 samples. It's important to note that during the base class learning phase, we exclude samples containing novel classes, and for novel class registration, we exclusively use samples that only contain novel classes. In our experiments, we also provide class name as texual information, so that we could effectively adapt pre-trained CLIP model to the GFSS setting. 

\noindent\textbf{GFSS Inference.}
During the inference stage, we test our model on the entire validation set of specified dataset (PASCAL-5$^i$ or COCO-20$^i$), which includes both base and novel classes simultaneously. Unlike in standard Few-Shot Segmentation (FSS), no additional information, such as support data, will be provided for each test sample.

\noindent\textbf{CIFSS Inference.}
For the evaluation of CIFSS setting, we evaluate the model on the classes belonging to base session, current and previous incremental novel sessions with the validation set. The performance of model with current novel session $S_n^t$ is evaluate on the test samples of $S_n^t$ and the previous sessions$\{S_n^1, \cdots, S_n^{t-1}\}$.

\section{Additional Results and Analysis for GFSS}
\noindent\textbf{Detailed Numerical Results for Each Fold.}
Table~\ref{pascal_fold} and Table~\ref{coco_fold} show the detailed  quantitative comparison for each fold in PASCAL-5$^i$ and COCO-20$^i$ datasets, respectively. It obvious that our approach outperforms other methods with significant improvement, which further demonstrates the effectiveness of the proposed method.
For example, our method attains \textit{h}IoU scores of 68.32, 63.11, 66.56, and 55.89 for each fold in PASCAL-5$^i$, significantly surpassing the performance of previous methods. Moreover, the considerable improvements across all folds of both PASCAL-5$^i$ and COCO-20$^i$ highlight our method's robustness against varying partitions of base and novel class sets.

\begin{table}[tp]
\centering
\caption{Comparison of the quantity of learnable parameters and inference speed during the 1-shot setting at the inference stage. Our model utilizes the pre-trained ViT-B/16, whereas the comparison methods employ ResNet-50 as their backbone.}
\label{table:params}
\setlength{\abovecaptionskip}{0cm}
\setlength{\belowcaptionskip}{-0.4cm}
\begin{threeparttable}
\scriptsize
\begin{tabular}{>{\arraybackslash}m{0.28\columnwidth} >{\centering\arraybackslash}m{0.28\columnwidth} >{\centering\arraybackslash}m{0.28\columnwidth}}
\toprule
 Method & \#Params & FPS \\ 
\midrule
 BAM~\cite{lang2022learning}      & 26.70M &  7.49  \\
 CAPL~\cite{tian2022generalized}     & 17.40M & 10.20   \\
 PKL~\cite{huang2023prototypical}      & 18.10M &  9.35 \\
 DIAM~\cite{hajimiri2023strong}     & 22.15M &  3.15  \\
 PoP~\cite{liu2023learning}      & 24.46M & 11.45   \\
 \rowcolor[gray]{.9} Ours     & 14.09M & 26.18   \\
\bottomrule
\end{tabular}
\end{threeparttable}
\end{table}

\noindent\textbf{Computation Efficiency.}
Table~\ref{table:params} presents a comparison of the number of learnable parameters and the inference speed (measured in FPS) during 1-shot inference with other methods. 
Despite being built on ViT-B/16, our method requires fewer learnable parameters and achieves a faster inference speed than those models using ResNet-50. Concerning learnable parameters, our model requires fewer than those of competing methods. This efficiency results from our approach of keeping the pre-trained model frozen while implementing an effective adaptation strategy for the GFSS task. The majority of learnable parameters are derived from the lightweight mask decoder and the visual prompts for the image decoder. In terms of inference speed, our model exhibits much faster inference speed (26.18 FPS) than previous methods. This further proves the effectiveness and efficiency of our model for the GFSS task.

\noindent\textbf{Effects of Different Backbones.}
Our model is uniquely crafted on the CLIP architecture using ViT-B/16, with specific design elements like visual prompts for the image encoder. Table~\ref{table:params} shows our ViT-based model's superior efficiency over models with a ResNet-50 backbone. However, to comprehensively evaluate its performance, we conducted an ablation study using ResNet-50 as the image encoder. As shown in Table~\ref{table:backbone}, with ResNet-50, our model still delivers impressive results, albeit with a roughly 2\% decrease in \textit{h}IoU. This drop could be attributed to the limited representational capacity of ResNet-50 and the absence of visual prompts to effectively adapt the pre-trained CLIP model for segmentation tasks. Considering efficiency and effectiveness, we finally choose the pre-trained CLIP with ViT-B/16 as the backbone.

\begin{table}[!t]
\Large
\setlength{\abovecaptionskip}{0cm}
\setlength{\belowcaptionskip}{-0.3cm}
\caption{\textbf{Effects of different backbones on PASCAL-5$^i$ and COCO-20$^i$}. We compare our model with ResNet-50 and ViT-B/16 as backbones.}
\begin{threeparttable}
\resizebox{1.0\linewidth}{!}{
\begin{tabular}{lcccccccc}
\toprule
\multirow{2}{*}{\bf Methods} & \multicolumn{4}{c}{\bf 1-shot} & \multicolumn{4}{c}{\bf 5-shot} \\ 
\cmidrule(l){2-5} \cmidrule(l){6-9}
& \bf $\emph{m}\text{IoU}_{\mathcal{B}}$ & \bf $\emph{m}\text{IoU}_{\mathcal{N}}$ & \bf $\emph{m}\text{IoU}_{\mathcal{O}}$ & \bf $\emph{h}\text{IoU}$ & \bf $\emph{m}\text{IoU}_{\mathcal{B}}$ & \bf $\emph{m}\text{IoU}_{\mathcal{N}}$ & \bf $\emph{m}\text{IoU}_{\mathcal{O}}$ & \bf  $\emph{h}\text{IoU}$ \\ 
\midrule
\multicolumn{9}{c}{\bf PASCAL-5$^i$} \\
\midrule
 \bf PPCN-Res50 & 73.47 & 48.15  & 68.34 & 57.96 & 73.66 &  57.95 & 71.32  & 64.53 \\ 
\bf PPCN-ViT  &  76.68 & 52.70 & 71.72 & 62.03 & 76.31 &  59.21 &  73.06  &  66.44 \\ 
\midrule
\multicolumn{9}{c}{\bf COCO-20$^i$} \\
\midrule
 \bf PPCN-Res50 & 48.36 & 25.17  & 42.76 & 33.15  & 48.25 & 31.33 & 44.85  & 37.64 \\ 
\rowcolor[gray]{.9} \bf PPCN-ViT & 48.82 & 26.79  & 43.83 & 34.59  & 49.05 & 31.52 & 45.17  & 38.34 \\ 
\bottomrule
\end{tabular}}
\end{threeparttable} 
\label{table:backbone}
\vspace{-4mm}
\end{table}

\begin{table*}[!th]
\setlength{\abovecaptionskip}{0cm}
\setlength{\belowcaptionskip}{0cm}
\caption{\textbf{Quantitative results under the 5-shot CIFSS setting on the PASCAL-5$^i$ and COCO-20$^i$}. Best-performing results are highlighted in bold. All baseline results come from \cite{huang2023prototypical}. Our model exhibits significant better performance than previous methods.}
\resizebox{\linewidth}{!}{
\begin{tabular}{c|l|c|ccc|ccc|ccc|ccc|ccc}
\toprule
\multirow{2}{*}{Datasets} & \multirow{2}{*}{Methods} & sesson 0 & \multicolumn{3}{c|}{session 1} & \multicolumn{3}{c|}{session 2} & \multicolumn{3}{c|}{session 3} & \multicolumn{3}{c|}{session 4} & \multicolumn{3}{c}{session 5} \\ \cline{3-18} 
& & $\emph{m}\text{IoU}_{\mathcal{B}}$ & $\emph{m}\text{IoU}_{\mathcal{B}}$ & $\emph{m}\text{IoU}_{\mathcal{N}}$ & $\emph{h}\text{IoU}$ & $\emph{m}\text{IoU}_{\mathcal{B}}$ & $\emph{m}\text{IoU}_{\mathcal{N}}$ & $\emph{h}\text{IoU}$ & $\emph{m}\text{IoU}_{\mathcal{B}}$ & $\emph{m}\text{IoU}_{\mathcal{N}}$ & $\emph{h}\text{IoU}$ & $\emph{m}\text{IoU}_{\mathcal{B}}$ & $\emph{m}\text{IoU}_{\mathcal{N}}$ & $\emph{h}\text{IoU}$ & $\emph{m}\text{IoU}_{\mathcal{B}}$ & $\emph{m}\text{IoU}_{\mathcal{N}}$ & $\emph{h}\text{IoU}$ \\ \hline
\multirow{5}{*}{PASCAL-5$^i$} & PFENet \cite{PFENet_TPAMI_2020} &74.43 &68.80 &18.82 &29.56 &66.61 &23.30 &34.52 &63.91 &17.77 &27.81 &60.85 &21.12 &31.36 &58.78 &15.65 &24.72\\
& iFS-RCNN \cite{nguyen2022ifs} &72.43 &70.49 &22.73 &34.38 &68.81 &25.11 &36.79 &67.74 &21.15 &32.24 &64.03 &25.50 &36.47 &62.42 &19.94 &30.22\\
& CAPL \cite{tian2022generalized} &74.86 &71.19 &22.83 &34.57 &68.88 &24.49 &36.13 &68.08 &24.07 &35.57 &64.89 &26.99 &38.12 &63.21 &20.84 &31.35  \\
& BAM \cite{lang2022learning} &75.83 &74.88 &24.41 &36.82 &70.91 &26.73 &38.82 &68.49 &25.55 &37.22 &65.36 &27.71 &38.92 &63.34 &22.58 &33.29 \\
& PIFS \cite{cermelli2021prototype} &75.04 &73.41 &23.39 &35.48 &68.82 &24.43 &36.06 &68.50 &26.76 &38.49 &66.61 &28.44 &39.86 &63.59 &26.63 &37.54 \\
& PKL \cite{huang2023prototypical}&75.49 &73.00 &30.64 &43.16 &71.52 &34.98 &46.98 &71.09 &33.06 &45.13 &70.37 &33.13 &45.05 &68.89 &34.56 &46.03 \\
\rowcolor[gray]{.9} \multicolumn{1}{c|}{\cellcolor{white}} &  \bf DPCN & \bf 79.39 & \bf 76.42 & \bf 49.94 & \bf 55.99 & \bf 76.19 & \bf 56.82 & \bf 64.62 & \bf 75.79 & \bf 60.83 & \bf 66.99 & \bf 75.81 & \bf 59.34 & \bf 66.54 & \bf 75.48 & \bf 56.67 & \bf 64.59 \\ \hline \hline
\multirow{5}{*}{COCO-20$^i$} & PFENet \cite{PFENet_TPAMI_2020} &54.11 &48.92 &8.81 &14.93 &44.30 &10.07 &16.41 &40.47 &6.64 &11.41 &39.95 &8.88 &14.53 &38.71 &9.90 &15.77\\
& iFS-RCNN \cite{nguyen2022ifs} &53.42 &52.28 &10.84 &17.96 &46.69 &12.11 &19.23 &43.77 &8.12 &13.70 &41.01 &10.10 &16.21 &40.06 &11.73 &18.15 \\
 & CAPL \cite{tian2022generalized} &54.43 &52.90 &11.69 &19.15 &47.18 &12.29 &19.50 &44.43 &8.86 &14.77 &41.39 &11.47 &17.9 &40.41 &12.22 &18.77 \\
 & BAM \cite{lang2022learning} &54.80 &52.73 &14.47 &22.71 &48.60 &14.43 &22.25 &46.72 &11.19 &18.06 &44.96 &14.08 &21.44 &43.35 &14.88 &22.16 \\
 & PIFS \cite{cermelli2021prototype} &54.27 &51.95 &14.68 &22.89 &48.04 &13.32 &20.86 &46.93 &11.82 &18.88 &45.08 &14.66 &22.12 &43.57 &14.13 &21.34 \\
 & PKL \cite{huang2023prototypical} &54.39
&51.82 &17.00 &25.60 &48.33 &16.65 &24.77 &47.10 &14.17 &21.79 &46.16 &16.67 &24.49 &45.22 &16.02 &23.66 \\
\rowcolor[gray]{.9} \multicolumn{1}{c|}{\cellcolor{white}} & \bf DPCN & \bf 55.99  &  50.50 & \bf 41.27 & \bf 45.66 & \bf 50.36 & \bf 42.00 & \bf 45.75 & \bf 50.25 & \bf 33.62 & \bf 40.28 & \bf 50.34 & \bf 33.47 & \bf 40.13 & \bf 49.78 & \bf 32.41 & \bf 39.19 \\
\bottomrule
\end{tabular}}
\label{table:CIFSS5shot}
\end{table*}

\begin{figure}
    \centering
    \includegraphics[width=0.85\linewidth]{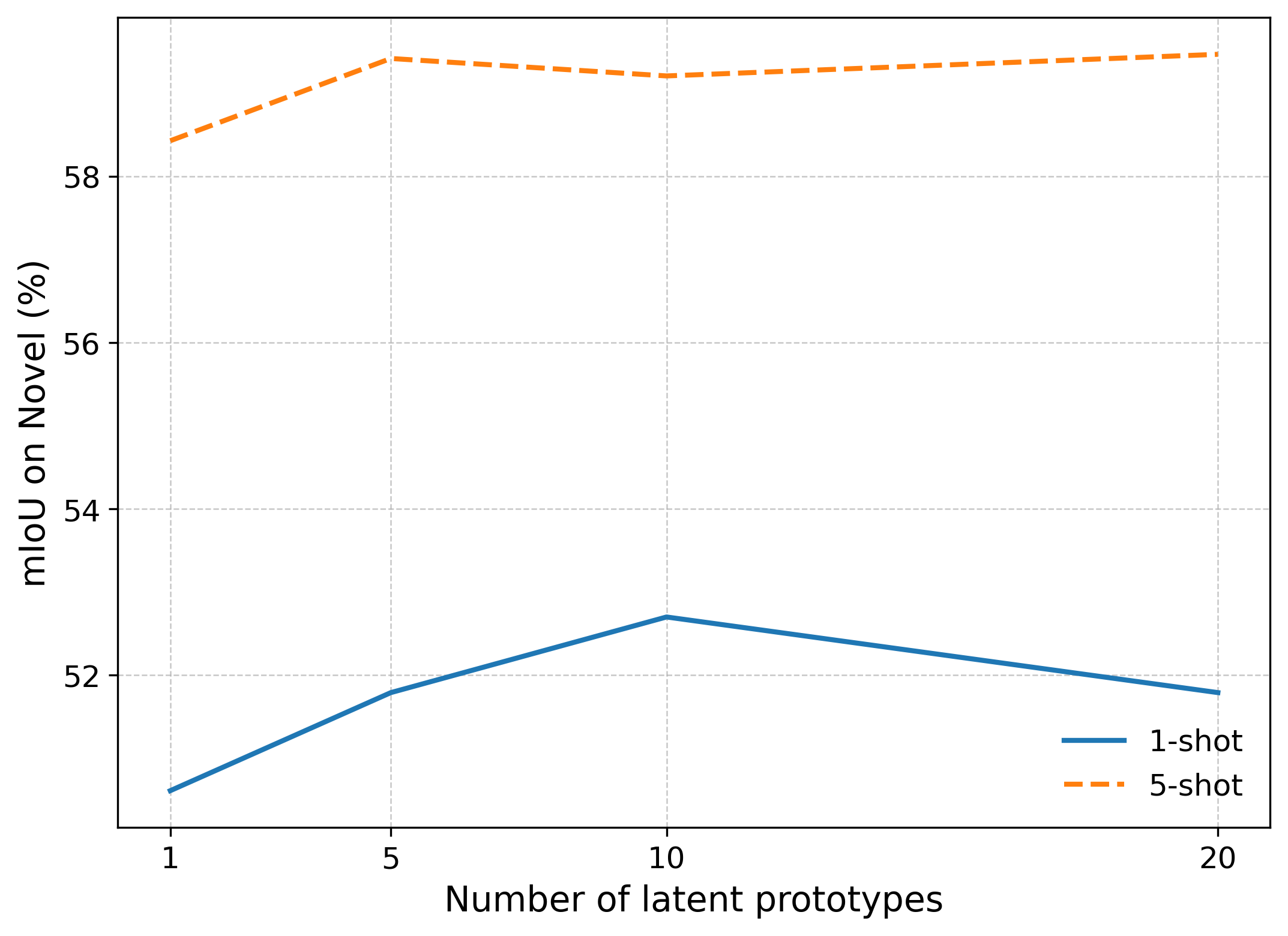}
    \caption{Effects of different number \(M\) of latent prototypes.}
    \label{fig:latent}
\end{figure}

\noindent\textbf{Effects of different number of latent prototypes.}
To analyze the impact of the number of sampled latent prototypes \(M\), we evaluate the model performance on novel classes by varying \(M\) while keeping other settings fixed.
As shown in Fig.~\ref{fig:latent}, increasing 
\(M\) initially improves performance, particularly in the 1-shot setting, as more latent prototypes provide a richer representation that captures intra-class variations.
However, when \(M\) becomes too large, performance slightly declines, suggesting that excessive latent prototypes may introduce noise or redundancy, making optimization more challenging.
This result highlights the importance of selecting an appropriate  \(M\)  to balance representation diversity and model stability in GFSS.

\section{Additional Results and Analysis for CIFSS}
The table~\ref{table:CIFSS5shot} presents our model's performance in class incremental few-shot semantic segmentation under the 5-shot setting. Consistent with the 1-shot scenario, our approach significantly outperforms recent methods in mIoU$_\mathcal{N}$ and \textit{h}IoU. The performance edge of our method grows with each additional novel session. For instance, in the fifth session, our method exceeds the previous state-of-the-art, PKL, by 18.46\% and 15.53\% on PASCAL-5$^i$ and COCO-20$^i$ respectively. These results confirm our method's capability in effectively handling segmentation for both base and novel classes in a dynamic, online context.

\begin{figure*}[!t]
\centering
\includegraphics[width=\linewidth]{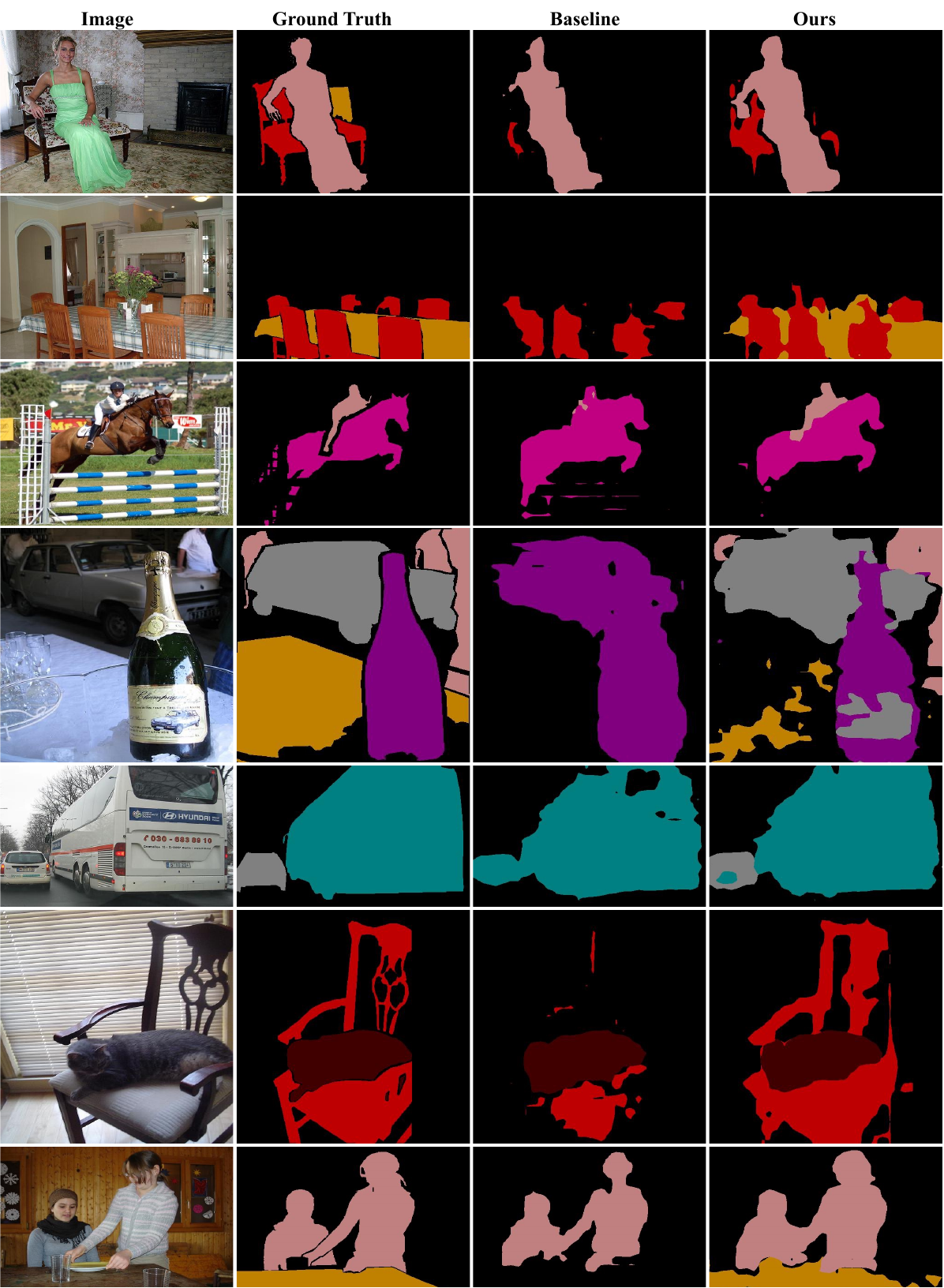}
\caption{\textbf{Qualitative examples of generalized few-shot semantic segmentation on PASCAL-5$^i$}. Results from the simple baseline and our method are reported. Examples with large object variations are exhibited.}
\label{fig:vis_pascal}
\end{figure*}
\begin{figure*}[!t]
\centering
\includegraphics[width=\linewidth]{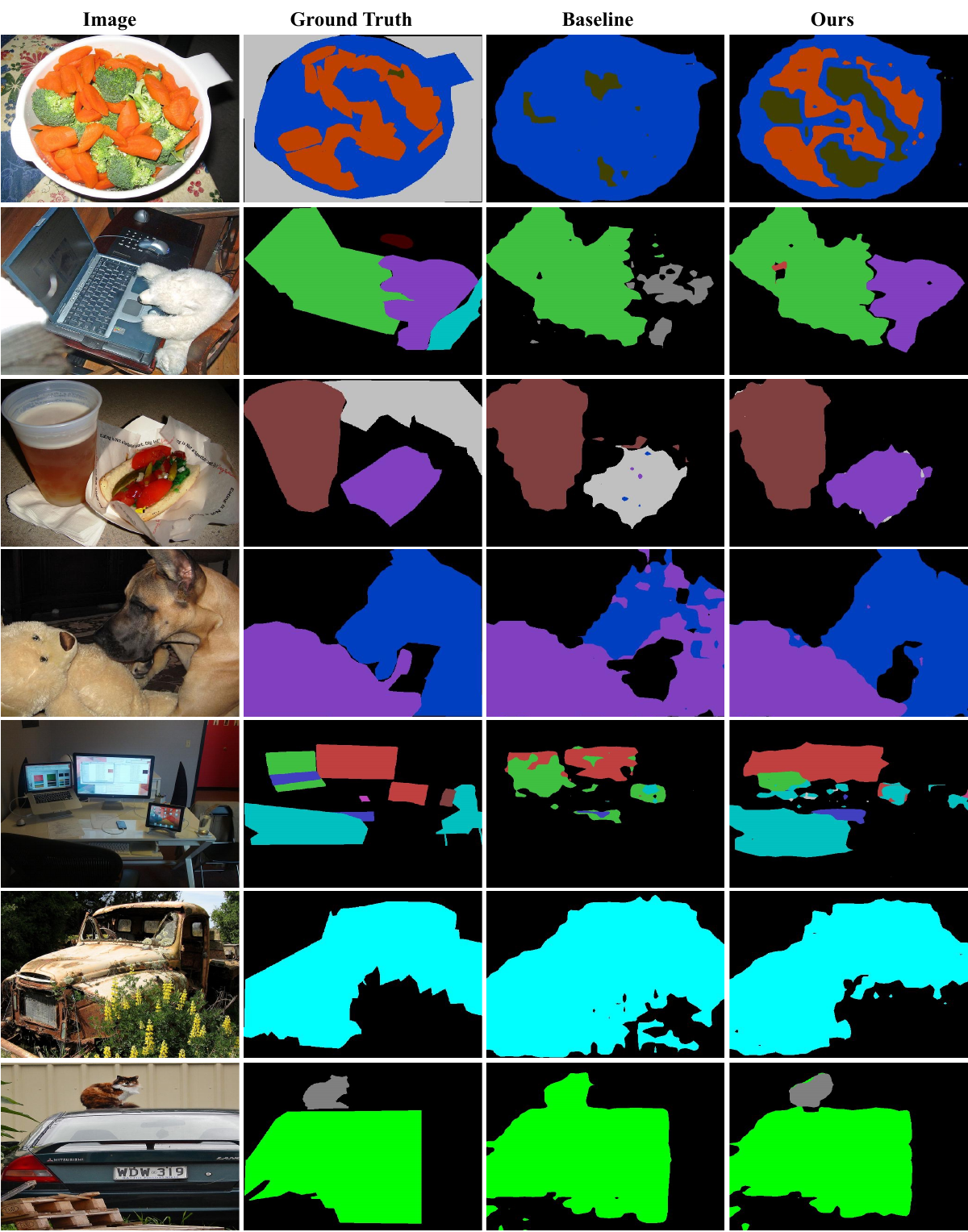}
\caption{\textbf{Qualitative examples of generalized few-shot semantic segmentation on COCO-20$^i$}. Results from the simple baseline and our method are reported. Examples with large object variations are exhibited.}
\label{fig:vis_coco}
\vspace{1.2cm}
\end{figure*}

\section{Additional Qualitative Results.}
In this section, we present additional qualitative results of our proposed method, showcasing its effectiveness in generalized few-shot semantic segmentation. Figure~\ref{fig:vis_pascal} illustrates that our method outperforms the baseline model on the PASCAL-5$^i$ dataset. For example, in the second row, while the baseline model fails to recognize the dining table, our model accurately distinguishes the pixels of both chairs and the dining table. Likewise, as depicted in Figure~\ref{fig:vis_coco}, our model demonstrates competitive performance on the challenging COCO-20$^i$ dataset compared to the baseline model, effectively handling significant object variations, such as scale and appearance. 


\end{document}